# Image Seam-Carving by Controlling Positional Distribution of Seams

Mahdi Ahmadi, Nader Karimi, Shadrokh Samavi
*Department of Electrical and Computer Engineering, Isfahan University of Technology,
Isfahan, Iran 84156-83111*

*Abstract*—Image retargeting is a new image processing task that renders the change of aspect ratio in images. One of the most famous image-retargeting algorithms is seam-carving. Although seam-carving is fast and straightforward, it usually distorts the images. In this paper, we introduce a new seam-carving algorithm that not only has the simplicity of the original seam-carving but also lacks the usual unwanted distortion existed in the original method. The positional distribution of seams is introduced. We show that the proposed method outperforms the original seam-carving in terms of retargeted image quality assessment and seam coagulation measures.

## I. INTRODUCTION

In recent years, by the increased diversity of screens and media display devices, the need to change the image and video aspect ratios has become more than before. The images can be resized to the new dimensions by scaling or cropping. However, these two simple methods do not consider the image content. On the other hand, in image retargeting algorithms, the image is mapped to the new dimensions so that its more essential contents are better preserved than its less essential contents.

One of the first attempts to image retargeting was the work introduced in [1] which is called seam-carving. In seam-carving, the pixel gradients are considered as their importance and the more critical pixels in this sense have more chance to be preserved. Although defining gradient as pixel importance seems naïve, seam-carving has shown to be very useful, specifically in low aspect ratio alterations. The authors of [1], also introduced an improved version of seam-carving in [2]. The new work tried to reduce the distortion in the retargeted image by taking into account the future gradient of image pixels, called forward energy.

Image retargeting methods vary in the modeling of the problem. Some of them model the problem by graph optimization [3], some model it by machine learning structures [4], and some model it by other optimization algorithms such as binary programming [5]. Also, the methods can be classified according to the image units that are removed or combined together. The units can be pixels [1], patches [6] or interpolated values between pixels [7]. The last group of methods are called continuous methods and usually perform a non-uniform scaling over the images. On the other hand, the pixel-based and patch-based methods are called discrete methods. Some methods are a combination of continuous and discrete methods that are iteratively performed to images. These methods are called multi-operator methods [8]. Finding the best method to retarget a specific image is very subjective; however, the researchers are trying to find methods that produce outputs plausible to more people on average.

An example of discrete methods is the work introduced in [3]. In this work, the output image is made by patches of the original image. The problem is modeled by a graph optimization problem that tries to find the relative shift of image patches. Another example is the work introduced in [9] that the patches of the original image are combined so that the translational symmetry of image content is preserved. The other examples are content-aware image cropping algorithms such as [10] and [11]. For example, the authors of [10] trained a convolutional neural network that predicts the crop area of the image based on object detection approaches. Their training is supervised; thus they developed a dataset of images by their crop areas in the form of bounding boxes.

In the continuous category, the non-uniform sampling is performed so that some similarity function on the original and retargeted images is maximized. In these methods, the more essential parts of the image are more sampled. Thus, the retargeted image has more pixels from such regions. The importance of image regions is defined differently in different papers. Nowadays, the saliency map of images has a determining roll in making importance maps for images. For instance, in the method introduced in [12], a system of equations is solved to find the new coordination of pixels in the video. The importance map is shaped by face detection, motion detection, and local saliency values. Another example is the work of [13] that is called the axis-aligned method. In this method, two energy terms are defined that one of them preserves rigidity, and the other preserves the similarity of the retargeted image. Then, by solving a quadratic optimization problem, the new coordinates for image pixels are found. The authors in [4], utilize two pre-trained neural networks that shape an end to end retargeting system. The self-supervised learning of the networks learns to produce an importance-map of the input images. Finally, a non-uniform sampling is performed to the image based on the produced saliency map.

Some image retargeting methods such as [14] and [15] are multi-operator methods. For example, the authors of [14] retarget the image by combining the three operators of seam-carving, cropping, and scaling. The image is retargeted in an iterative manner and at every iteration, one of the operators is selected to optimize an energy function. In [15] the operators are seam-carving, cropping, warping, and scaling. The sequence of operators is determined by optimizing a function called perceptual similarity measure. In [16-17], multiple operators such as carving and scaling are used. In [18], semantic segmentation is applied before carving less essential regions of the image.

In this paper, we modify the seam-carving algorithm presented in [2] so that the retargeted image is less distorted than the work in [2]. The proposed method does not require resources more than the original seam-carving. The only extra process is a suitable energy function preparation for the original method. While the proposed method is as simple as the original seam-carving, it has better performance in terms of Retargeted Image Quality Assessment (RIQA) measures. However, the proposed method cannot compete with the state-of-the-art retargeting methods. The reason is that the state-of-the-art mainly require preprocessing or post-processing stages or utilize neural networks, which can be the bottleneck of the whole task. For example, image saliency detection is a typical preprocessing stage that is a complex task itself.

We explain the proposed seam-carving method in the next section. The results of the work are analyzed and are compared to the original seam-carving method in section III. Finally, we conclude the work and discuss possible future works in section IV.

## II. THE PROPOSED METHOD

Since the proposed method is based on the improved seam-carving method presented in [2], we first explain the seam-carving method and then our modifications to it. The improved seam-carving process can be modeled in two ways: dynamic programming and graph-cut. The graph-cut model is more general and can also be used in video retargeting, and so is selected here. The objective is to find 8-connected passes of pixels called seams that will be deleted. The method can also be used for image enlargement so that new pixels are inserted next to the location of seams. In this paper, we change the width of images, and so we have vertical seams. Since the horizontal seam-carving is similar to vertical seam-carving, we do not explain it.

In seam-carving, the image is retargeted iteratively by deleting vertical seams at each iteration. The goal in each iteration is to find seams that have the least energy. In its graph, the representation of the seam-carving method, the image pixels are nodes of a directed graph. To find the best seam to be deleted, the graph-cut algorithm is used in each iteration. Graph-cut is a mathematical tool to cut a graph so that the removed arcs have the least (or greatest) summation of weighs among all other possible cuts.

In directed graphs, the directed arcs originate from a source node and go to a target node. In the seam-carving, a source node and a target node are added to the graph. Each arc stands for the energy of its beginning node. Hence, cutting an arc means removing the pixel/node at its beginning. Note that in directed graph-cut, all removed arcs must be directed from source to the target node. Therefore, cutting an arc that is in the opposite direction does not count in energy calculation. This fact enables us to design the graph in a way that satisfies the problem requirements. For example, in Fig. 1.a, the energy of node $p_{i,j}$ is $E_1$ and the infinity arcs are added to preserve the continuity and monotonic nature of seams that are required in the seam-carving algorithm. Figure 1.a is the graph representation of [1], which was the previous work of the authors of [2]. But in their new work, i.e. [2], the individual pixels normally do not have an energy term but can be added as an extra energy term.

Authors of [2] put the source and target nodes at the left and right of image nodes/pixels respectively and connect all the nodes of the left column of the image to source and all nodes of the right column to the target node. The energy terms $+LR, -LU$ and $+LU$ in Fig. 1.b are the forward energy that means they will affect the image *after* removing the pixels. For example, if the arc with energy $+LR$ is cut, $+LR$ amount of energy will be added to the image at this place. This energy is added because of the pixels $p_{i,j+1}$ and $p_{i,j-1}$ become neighbors after the pixel $p_{i,j}$ is deleted. The added energy value is the gradient of these two pixels i.e. $p_{i,j+1}$ and $p_{i,j-1}$ and is obtained by the following equation:

$$+LR = |I(i, j + 1) - I(i, j - 1)| \quad (1)$$

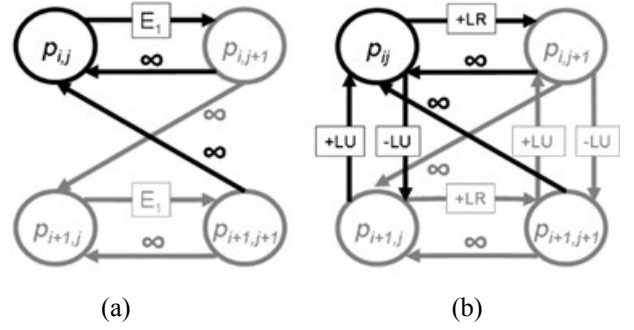

Fig. 1. The state diagram for (a) method in [1], (b) method in [2].

In this equation, $I$ represents the gray value of image pixels. Similarly $-LU$ and $+LU$ are obtained by:

$$+LU = |I(i - 1, j) - I(i, j - 1)| \quad (2)$$
$$-LU = |I(i + 1, j) - I(i, j - 1)| \quad (3)$$

In this paper an energy term $E_{ramp}$ is added to $+LR$ that regularly controls the distribution of seams. By adding $E_{ramp}$ in this energy term, the seams are so distributed that there will be less coagulation in seams. The lower density of seams means that the seams are distributed in more image regions and are not placed next to each other. This term is the energy allocated to the pixel $p_{i,j}$ and is calculated by the equation (4):

$$E_{ramp}(i,j) = \begin{cases} \alpha \ (j \bmod R_1) & if \ (i \bmod R_2) = 0 \\ 0 & otherwise \end{cases} \quad (4)$$

In this equation, $i$ is the pixel row, $j$ is pixel column, mod is the remainder operator, $\alpha$ is the energy magnitude constant, and $R_1$ and $R_2$ are two parameters that describe the energy

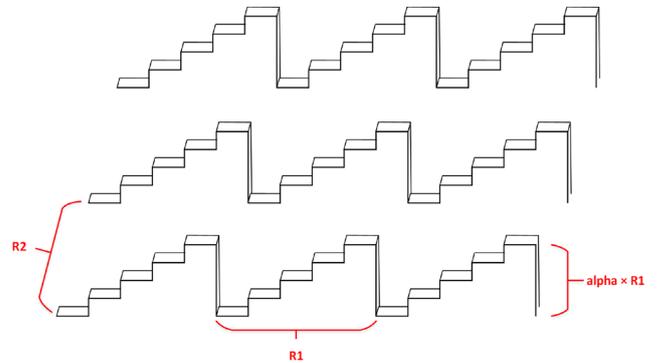

Fig. 2. The illustration of $E_{ramp}$ in the proposed method.

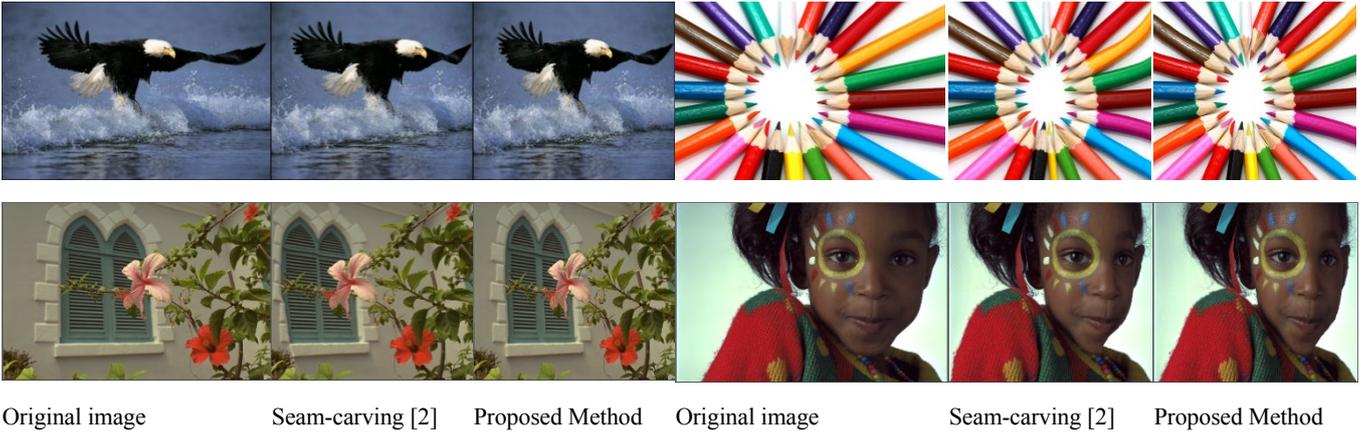

Original image      Seam-carving [2]    Proposed Method    Original image      Seam-carving [2]    Proposed Method

Fig. 3. Comparison of the proposed method to the original seam-carving [2]. Images are retargeted by 25%.

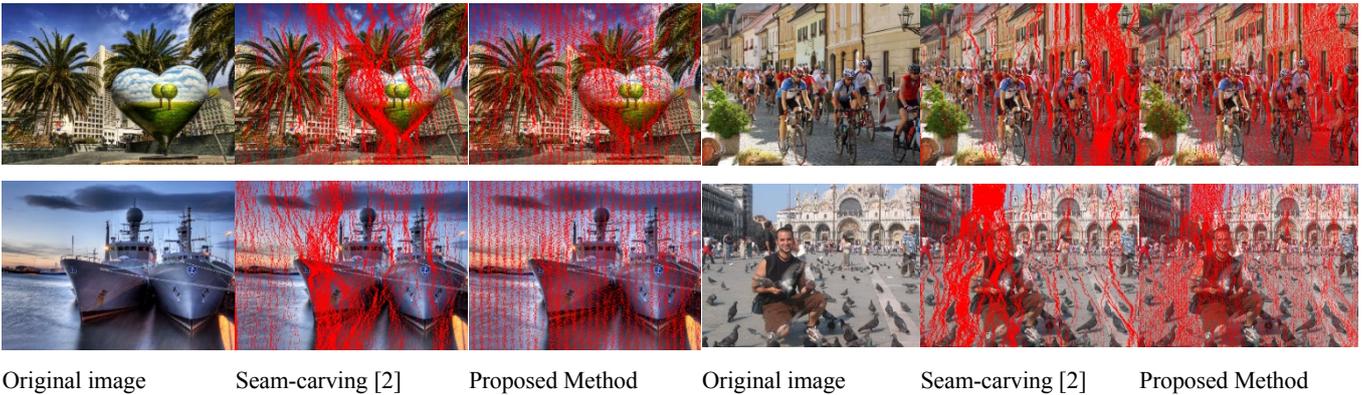

Original image      Seam-carving [2]    Proposed Method    Original image      Seam-carving [2]    Proposed Method

Fig. 4. Seam distribution in the seam-carving method [2] and the proposed method. Images are retargeted by 25%.

ramps. The distance of energy ramps in different rows is determined by $R_2$ and the maximum steps of the ramp is determined by $R_1$. Thus, the maximum magnitude of ramps is $\alpha R_1$ as can be seen in Fig. 2.

By this new energy term, not only the forward energy is considered, but also the distribution of seams is more controlled. Also, by tuning $\alpha$, the contribution of the forward-energy and seam controlling can be adjusted. Generally, the seams cross the lower values of ramps unless there is a more powerful forward energy in the place. By increasing the $R_2$ parameter, the seams have more freedom in between the ramp rows. It is obvious that there is no randomness in this method, and the ramps are located on a regular basis. Also, the parameters are set without taking the image features into account.

### III. EXPERIMENTS

The proposed method is implemented in MATLAB. Two datasets of CUHK [e] and RetargetMe [f] are utilized for the analysis and comparison. These two datasets are the benchmark datasets for image retargeting quality assessment and are commonly used in the literature [19-20]. CUHK consists of 57 images, and RetargetMe consists of 80 images that are from different categories such as the face, symmetry, and so on. We set the parameters of $\alpha, R_1$, and $R_2$ to 0.0625, 4, and 5.

The proposed method is compared to the original seam-carving [2] by two criteria. One criterion is called Seam Coagulation Measure (SCM), which is explained in the following parts. The other criterion is Aspect Ratio Similarity (ARS) that is introduced in [21] as a RIQA criterion. We compare the results on both CUHK and RetargetMe datasets and show the superiority of the proposed method.

Before evaluating the proposed method by objective measures, i.e., ARS and SCM, we take a glance at the outputs. Some retargeted images by the seam-carving process of [2] and the proposed method are shown in Fig. 3. As seen in this figure, the retargeted images obtained by the proposed method are less distorted, and the shapes of objects are preserved better. The removed seams for some images are shown in Fig. 4. We can see in Fig. 4 that in the proposed method, seams are removed in a way that less contiguous seams are removed. Distribution of extracted seams enhances the visual quality of the resulting image because there are fewer discontinuities as compared with the original seam-carving algorithm.

To evaluate the ability of the proposed method in distributing the seams, we introduce the SCM measure. SCM shows how the algorithm prevents the coagulation of seams. Higher SCM means that the retargeted image suffers more seam coagulation and thus would be more distorted in terms of pixel discontinuity. The following equation obtains this criterion:

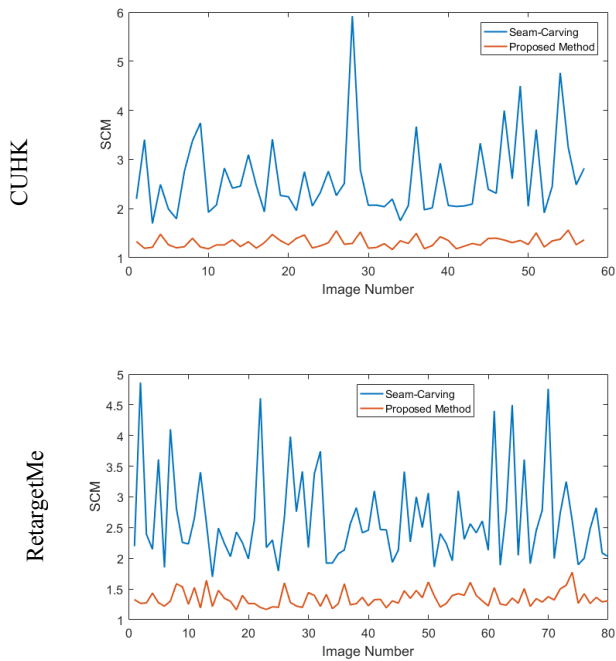

Fig. 5: SCM values of seam-carving [2] and the proposed method. First row: the results of CUHK dataset images, second row: the results of the RetargetMe dataset.

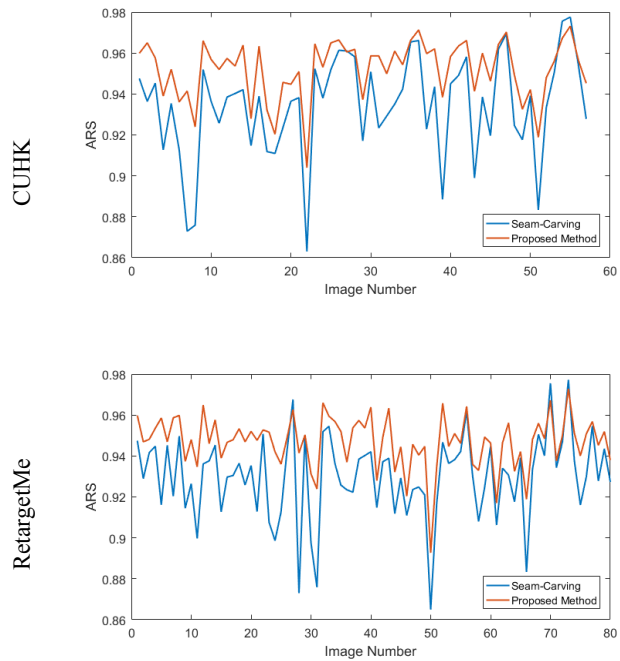

Fig. 6: SCM values of seam-carving [2] and the proposed method. First row: the results of CUHK dataset images, second row: the results of the RetargetMe dataset.

$$SCM = \sqrt{\frac{\sum_{h=1}^{H} \sum_i L_i^2}{H \times (W - W')}} \qquad (5)$$

In this equation, $H$ and $W$ are the original image height and width, respectively, and $W'$ is the width of the retargeted image. $L_i$ is the length of $i^{th}$ line segment in the $h^{th}$ image row. It is defined as the number of connected seam pixels. For example, if $i^{th}$ line segment contains 4 pixels, then $L_i = 4$.

The SCM measure for the images of both CUHK and RetargetMe are illustrated in Fig. 5. As can be seen in this figure, the retargeted images in the proposed method are less distorted in terms of SCM measure.

The other criterion that is used to compare the proposed method to the original seam-carving is ARS [21] that is one of the known RIQA methods. In this criterion, the higher ARS values represent better quality of the retargeted image. The ARS values for the images of CUHK and RetargetMe datasets are shown in Fig. 6. According to this figure, the proposed method is again better than the seam-carving process in terms of ARS.

IV. CONCLUSION

In this paper, a new seam-carving algorithm was introduced. The algorithm reduced the distortions existing in the original seam-carving by regularly controlling of seams distribution. The proposed method did not require extensive pre/post-process, such as saliency detection, and is obtained by modifying the energy function of the original seam-carving method. The results showed that the proposed method outperforms the original seam-carving in terms of the quality of the retargeted images.

For future work, we can put the energy terms in locations that are determined by image local features. Also, their energy and length can be tuned based on image features. So, the seams distribution may be better controlled according to the images. Another possible future work would be generalizing the proposed method for video retargeting. For video seam carving, instead of removing seams, 3D manifolds will be removed from the streaming video frames.